\documentclass[lettersize,journal]{IEEEtran}
\usepackage{amsmath,amsfonts}
\usepackage{algorithmic}
\usepackage{algorithm}
\usepackage{array}
\usepackage[caption=false,font=normalsize,labelfont=sf,textfont=sf]{subfig}
\usepackage{textcomp}
\usepackage{stfloats}
\usepackage{url}
\usepackage{verbatim}
\usepackage{graphicx}
\usepackage{cite}

\usepackage{multirow}
\usepackage[colorlinks,linkcolor=blue]{hyperref}
\usepackage{enumerate}
\usepackage{stmaryrd}
\usepackage{bigstrut}
\usepackage{ulem}
\usepackage{gensymb}
\usepackage[scaled=1.0]{helvet}
\usepackage{tabularx}
\usepackage{soul}
\soulregister\cite7
\soulregister\eqref7
\soulregister\ref7

\usepackage{threeparttable}

\newcommand\norm[1]{\left\lVert#1\right\rVert}

\begin{document}

	\title{ALIKE: Accurate and Lightweight Keypoint Detection and Descriptor Extraction}

	\author{Xiaoming~Zhao,
		Xingming~Wu,
		Jinyu~Miao,
		Weihai~Chen*,~\IEEEmembership{Member,~IEEE},		
		Peter~C.~Y.~Chen,
		and~Zhengguo~Li*,~\IEEEmembership{Senior~Member,~IEEE}
		\thanks{This work was supported by the National Nature Science Foundation of China under Grant No. 61620106012. (\textit{Corresponding authors: Weihai Chen; Zhengguo Li.})}
		\thanks{Xiaoming Zhao, Xingming Wu, Jinyu Miao, and Weihai Chen are with the School of Automation Science and Electrical Engineering, Beihang University, Beijing, 100191 (e-mail: xmzhao@buaa.edu.cn, wxmbuaa@163.com, mjy0519@buaa.edu.cn, and whchen@buaa.edu.cn).}
		\thanks{Peter C. Y. Chen is with the Department of Mechanical Engineering, National University of Singapore, Singapore (email: mpechenp@nus.edu.sg).}
		\thanks{Zhengguo Li is with the SRO department, Institute for Infocomm Research, 1 Fusionopolis Way, Singapore (email: ezgli@i2r.a-star.edu.sg).}
	}
	
	\maketitle
	
	\begin{abstract}
		Existing methods detect the keypoints in a non-differentiable way, therefore they can not directly optimize the position of keypoints through back-propagation. To address this issue, we present a partially differentiable keypoint detection module, which outputs accurate sub-pixel keypoints. The reprojection loss is then proposed to directly optimize these sub-pixel keypoints, and the dispersity peak loss is presented for accurate keypoints regularization. We also extract the descriptors in a sub-pixel way, and they are trained with the stable neural reprojection error loss. Moreover, a lightweight network is designed for keypoint detection and descriptor extraction, which can run at 95 frames per second for $640\times 480$ images on a commercial GPU. On homography estimation, camera pose estimation, and visual (re-)localization tasks, the proposed method achieves equivalent performance with the state-of-the-art approaches, while greatly reduces the inference time.
		
	\end{abstract}
	
	\begin{IEEEkeywords}
		Keypoint detection, keypoint descriptor, deep learning, local feature, image feature extraction, image matching
	\end{IEEEkeywords}

	\IEEEpeerreviewmaketitle
	
	\section{Introduction}
	
	\begin{figure}[!t]
		\centering
		\includegraphics[]{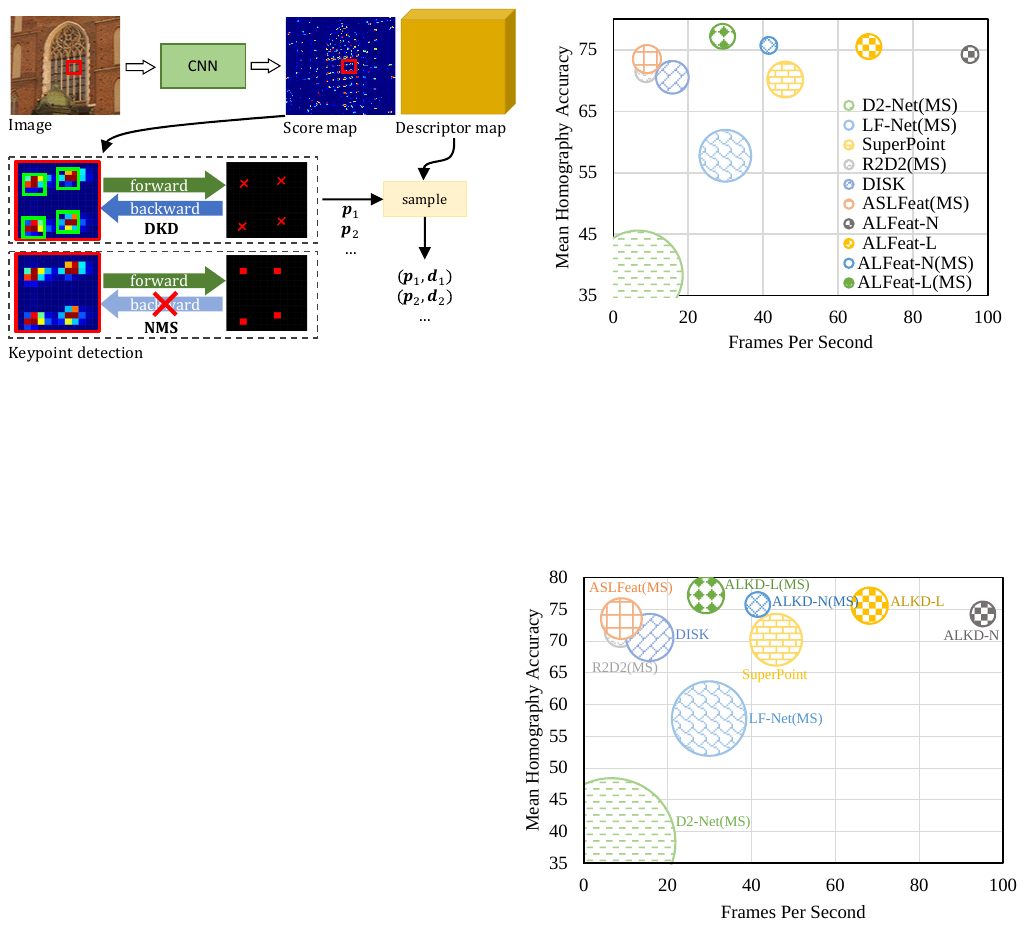}
		\caption{We propose a Differentiable Keypoint Detection (DKD) for score map based keypoint detection and descriptor extraction. The keypoint detection on a score map patch (the red rectangle area) is illustrated in this figure. Compared with the Non-Maximum-Suppression (NMS) based methods, the DKD can back-propagate the gradients and produce sub-pixel keypoints. Thus we can directly optimize the position of detected keypoints to be accurate.}
		\label{fig_1}
	\end{figure}

	\IEEEPARstart{S}{parse} keypoints and descriptors are compact representations for efficient image matching \cite{shotmatching}. Hence they are widely used in real-time visual applications like simultaneous localization and mapping (SLAM) systems \cite{orbslam2,slamdyn}, and high dynamic range imaging (HDRI) \cite{kou2017intelligent,zheng2013hybrid}.

	For keypoint detection and descriptor extraction, early handcrafted algorithms \cite{sift,orb,rgbdkeypoint} were built upon limited human heuristics, which could lead to unstable keypoints and confusable descriptors in complex images. So neural networks are explored for this task in recent years. Early neural network based methods mainly focus on descriptor extraction from image patches \cite{l2net,hardnet,sosnet}. Latter, many excellent methods address the keypoint detection and descriptor extraction with a single network \cite{superpoint,r2d2,d2net,aslfeat}. Some of them \cite{lift,lfnet,tilde,quadnet,superpoint,r2d2,sekd} treat the keypoint detection as a score map estimation problem, where the score of a pixel indicates the probability that it is a keypoint. And then they train the intermediate score map with the synthetic ground truth score map and/or the similarity of score patches. Others \cite{d2net,aslfeat,d2d,ur2kid} define the score map based on the spatial and/or channel variation of dense feature map.
	
	However, both of them have to detect the keypoints on score map with Non-Maximum-Suppression (NMS). The NMS simply chooses the pixel with maximum local score, it is non-differentiable and the gradients can not backward though (left bottom in Fig.\ref{fig_1}). So the position of detected keypoints can not be optimized directly. We observed that a keypoint is supported by its local score patch (the green areas in the middle score patch of Fig.\ref{fig_1}). The score distribution in the patch would influence the keypoint position, even if the position of maximum score remains unchanged. This allows us to accurately capture subtle changes of local score distribution. To this purpose, we adopt the score map based pipeline (Fig.\ref{fig_1}) for Differentiable Keypoint Detection (DKD). The DKD applies the \texttt{softargmax} \cite{softargmax,sun2018integral,gu2021removing} operation on local score patches. Thus the gradients can backward from keypoints to score map (middle row in Fig.\ref{fig_1}). We present the keypoint reprojection loss to train the sub-pixel keypoints detected from DKD, which directly minimizes the reprojection distance of detected keypoints between images. Inspired by the peak loss \cite{r2d2,sekd}, a dispersity peak loss is further introduced to avoid blob scores on score map, which forces the score map to be accurately ``peaky'' at the keypoint position.
	
	Besides the sub-pixel keypoints from score map, the descriptors are also sampled in a sub-pixel way from the dense descriptor map (Fig.\ref{fig_1}). The widely use method to train sparse descriptors is the triplet loss \cite{superpoint,d2net,aslfeat,sekd}, but our experiments indicate that the training of sub-pixel sparse descriptors with triplet loss is tricky and unstable, as they only cover the sampled keypoints of entire descriptor map. Inspired by sparse-to-dense matching methods \cite{s2dnet0,s2dnet,wang_learning_2020, nre}, we adopt the recent proposed neural reprojection error (NRE) loss \cite{nre}, which covers the entire descriptor map in training and thus provides more stable convergence.
	
	On the other hand, estimation of score map and descriptor map must be very efficient, as the keypoint detection and descriptor extraction is a fundamental task for many real-time applications. However, existing methods pay more attention on the matching performance rather than running efficiency. To further improve the running efficiency for robots \cite{droge2021dual}, a lightweight convolutional neural network (CNN) is designed by concatenating multi-level features \cite{aslfeat,sekd,mlifeat}, for both localization accuracy and representation capabilities. The experiments indicate that this lightweight network has comparable performance to existing methods while runs much faster.

	To summarize, the main contributions of this paper are as follows:
	\begin{itemize}
		\item We present a differentiable keypoint detection module, as well as the reprojection and dispersity peak loss for accurate and repeatable keypoints training.	
		\item We employ the NRE \cite{nre} loss to train the estimated dense descriptor map so that the model can converge more stably than using the triplet loss.
		\item A lightweight network aggregating hierarchical features is designed for efficient keypoint detection and descriptor extraction, which can run at 95 FPS (frame per second) on a commercial GPU while achieving comparable performances to state-of-the-art (SOTA) approaches.	
	\end{itemize}
	
	The rest of this paper is organized as follows. Section \ref{sec_related} reviews the deep learning based methods. Section \ref{sec_method} introduces the lightweight network, proposes the differentiable keypoint detection module, and presents the training losses. In section \ref{sec_exp}, we first analyze each part of the proposed method, then conduct the evaluation and discussion on different tasks. Finally, a conclusion is given in section \ref{sec_con}.

	\section{Related Works}
	\label{sec_related}
	Deep learning based methods can be roughly divided into three categories: the patch-based, score map based and the description-and-detection methods.

	\begin{figure*}[!t]
		\centering
		\includegraphics[]{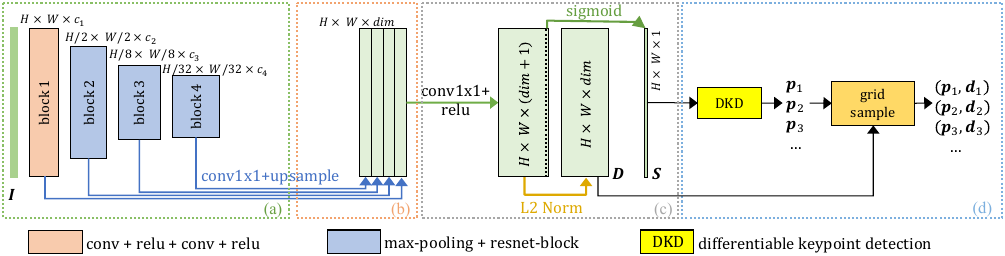}
		\caption{The proposed network architecture. (a) image feature encoder, (b) feature aggregation, (c) feature extraction head, (d) differentiable keypoint detection and descriptor sampling. The $\boldsymbol{I}$ denotes the input image, $\boldsymbol{p}$ is the position of keypoint, and $\boldsymbol{d}$ represents the descriptor of a keypoint.}
		\label{fig_accnet}
	\end{figure*}

	\subsection{Patch-based methods}
	The early patch-based methods only extract descriptors from image patches. The Matchnet \cite{matchnet} estimates similarity of descriptors, and trains them with cross entropy loss. Latter, TFeat \cite{tfeat} introduces the triplet loss for patch descriptors. And it is then widely used in latter patch-based methods \cite{l2net,lift,hardnet,sosnet}. L2-Net \cite{l2net} presents a progressive sampling strategy for triplet sampling. LIFT \cite{lift} mimics the SIFT \cite{sift}. HardNet \cite{hardnet} and SOSNet \cite{sosnet} introduce the hardest negative triplet and second order similarity of descriptors. However, the patch-based methods only focus on descriptors extraction, and their receptive field is limited in the image patch.
	
	\subsection{Score map based methods}
	These methods estimate a score map and a descriptor map, where the score map indicates keypoint probability. Tilde \cite{tilde} first trains the score map on webcam dataset with SIFT \cite{sift} keypoints as ground truth. Quad-networks \cite{quadnet} trains the score map by ranking scores to eliminate the need of ground truth labeling. And KeyNet \cite{keynet} estimates the score map with the handcrafted and learned features, it extracts keypoints from score map with \texttt{softargmax} \cite{softargmax}. Besides pure keypoint detection, recent methods estimate both the score map and descriptor map. LFNet \cite{lfnet} also uses \texttt{softargmax} \cite{softargmax}, but it still trains on score map rather than keypoints. SuperPoint \cite{superpoint} first trains a MagicPoint model on synthetic dataset, and then bootstraps the score map on real images with homographic adaption strategy, and its descriptors are trained with triplet loss. This strategy is also adopted in MLIFeat \cite{mlifeat} and SEKD \cite{sekd}. R2D2 \cite{r2d2} identifies keypoints as reliable and repeatable positions in image and trains reliability through AP loss \cite{aploss}. HDD-Net \cite{hddnet} weights the features with \texttt{softargmax} scores in grids to train the score map and feature map simultaneously. Furthermore, DISK \cite{disk} and reinforced SP \cite{bhowmik_reinforced_2020} relax the keypoint detection and descriptor matching as probabilistic processes and train the network with reinforcement learning.
	
	However, all these methods except KeyNet \cite{keynet} train on intermediate score map rather than directly on keypoints, as they are extracted with non-differentiable NMS. Our DKD is most similar to KeyNet \cite{keynet} and also utilizes \texttt{softargmax}. But our method dose not require handcrafted features and any pseudo keypoint annotations. KeyNet \cite{keynet} detects keypoints on fixed patches and can not handle the keypoints on patch boundaries, whereas our keypoints are extracted from flexible potential positions and therefore no boundary issues. Besides keypoints training, previous works mainly adopt the triplet loss to train descriptors, which is proved to be unstable in our experiments for sub-pixel descriptors. Inspired by sparse-to-dense matching \cite{s2dnet,wang_learning_2020, nre}, we utilize the NRE loss \cite{nre} to train the sub-pixel descriptors.

	\subsection{Description-and-detection methods}
	Unlike score map based methods (detection-then-description), description-and-detection methods recognize keypoints as distinctive positions in an image and generate the score map by computing distinctiveness of the descriptor map or feature maps. D2Net \cite{d2net} first proposes this concept, it applies channel-wise ratio-to-max and spatial-wise softmax on the descriptor map to compute score map. ASLFeat \cite{aslfeat} improves it with channel-wise and spatial-wise peakiness on multi-layer feature maps. UR2KiD \cite{ur2kid} computes the score map with the L2 response of features. And D2D \cite{d2d} selects the absolute and relative salient points from feature map as keypoints.
	However, they also have to supervise on the computed score map rather than directly on the keypoints, as they also use the no-differentiable NMS to detect keypoints.

	\section{Methods}
	\label{sec_method}
	
	Following the score map based approaches \cite{superpoint,r2d2}, for an input image $\boldsymbol I \in \mathbb{R}^{H\times W\times 3} $, the network first estimates a score map $\boldsymbol S  \in \mathbb{R}^{H\times W}$ and a descriptor map $\boldsymbol D  \in \mathbb{R}^{H\times W \times dim} $. Then sub-pixel keypoints $\{\boldsymbol{p}=[u,v]^T\}$ are detected with the DKD from the score map $\boldsymbol S$, and their corresponding descriptors $\{\boldsymbol{d}\in \mathbb{R}^{dim}\}$ are sampled from $\boldsymbol D$. 
	In the following, we first introduce the network architecture and the DKD module. Then the training losses for accurate keypoints and discriminative descriptors are presented.

	\subsection{Network architecture}
	\label{accnet}		
	
	As illustrated in Fig.\ref{fig_accnet}, the network is designed to be as lightweight as possible to improve running efficiency \cite{tonemapmatching}. It only has a basic encoder for feature extraction. Then the feature aggregation module assembles multi-level features \cite{aslfeat,sekd,mlifeat} to retain both the localization and representation capabilities. For accurate localization performance, the feature extraction head estimates the score map $\boldsymbol 
	S$ and descriptor map $\boldsymbol D$ under original image resolution. The details for each part are as follows:
	\begin{enumerate}[(a)] 
		\item \textbf{The image feature encoder} encodes the input image $\boldsymbol{I} \in \mathbb{R}^{H\times W\times 3} $ to feature maps. It contains four blocks. The first block is a two-layer $3\times 3$ convolution with ``ReLU'' activation \cite{relu}, and the last three blocks contain a max-pooling layer and a $3\times 3$ basic ResNet block \cite{resnet}. The number of output features for $i$-th module is denote as $c_i$. The downsample rate of max-pooling in block 2 is $1/2$, and $1/4$ in block 3 and 4. Under this formulation, the maximum receptive field is $204\times 204$ on the image.
		
		\item \textbf{The feature aggregation module} aggregates multi-level features from the encoder. An $1\times 1$ convolution and bilinear up-sampling are first used to adapt the channels, and then they are simply concatenated together. 
		
		\item \textbf{The feature extraction head} outputs an $H\times W \times (dim+1)$ feature map in which the first $dim$ channels are L2 normalized as the descriptor map $\boldsymbol D$ and the last channel is normalized by ``Sigmoid'' activation as score map $\boldsymbol S$. 
		
		\item \textbf{The differentiable keypoint detection and descriptor sampling} first detects the sub-pixel keypoints from the score map $\boldsymbol S$ (section \ref{sec_dkd}), and then samples their descriptors from the dense descriptor map $\boldsymbol D$. 
		
	\end{enumerate}

	\begin{figure*}[!t]
		\centering
		\includegraphics[]{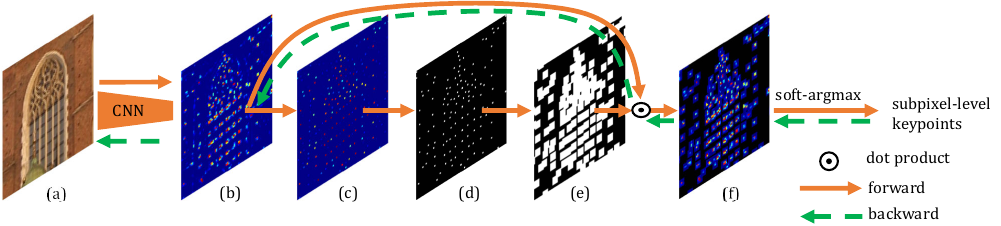}
		\caption{The differentiable keypoint detection (DKD) module. (a) input image, (b) the estimated score map, (c) the NMSed score map, (d) the coordinates after threshold, (e) the local windows for differentiable keypoint detection, (f) the scores in local windows. It first extracts the pixel-level keypoints by using NMS and threshold, then the scores in local windows are used to extract the subpixel-level keypoints. The gradient flows in the backward process are marked as dash green arrows. And all the score maps are encoded with ``jet'' colormap. Best viewed in color.}
		\label{fig_softdetect}
	\end{figure*}
	
	\subsection{Differentiable keypoint detection module}
	\label{sec_dkd}

	To detect keypoints in score map $\boldsymbol S$, a widely used method is the NMS \cite{superpoint,r2d2,d2net}. It finds the pixels with the maximum score within local windows. This operation is equivalent to the \texttt{argmax} in a local $N\times N$ window
	\begin{equation} \small
		\label{equ_argmax}
		[\hat i, \hat j]_{NMS}^T = \arg \max_{i,j} \left\{ s(i,j) \ |\ 0\le i,j < N \right\},
	\end{equation}
	where $s(i,j)$ denotes the score of the local position $(i,j)$. However, in this formula, the output position is decoupled with the score map, thus it is a non-differentiable operation and can not embrace the power of deep learning.
	
	To couple the keypoints with score map, we propose extracting differentiable keypoints from local windows with \texttt{softargmax}.
	Formally, the NMSed score map (Fig. \ref{fig_softdetect}(c)) is first obtained by suppressing the non-maximum scores $s=\boldsymbol S(u,v)$ in local $N\times N$ windows:
	\begin{equation} \small
		s = \begin{cases}
			s_{max} & s=s_{max}\\
			0 & others
		\end{cases}, 
	\end{equation}
	where $s_{max} = \max s(i,j)$ is the local maximum score. Then, a threshold $th$ is applied on the NMSed score map to filter out low response scores (Fig. \ref{fig_softdetect}(d)). In NMS-based methods, keypoints $\{[u,v]_{NMS}^T\}$ are extracted in this step. We step further by peeking the scores in the local windows centered on NMS keypoints $\{[u,v]_{NMS}^T\}$ (Fig. \ref{fig_softdetect}(f)), and extracting local soft coordinates $\{[\hat i, \hat j]_{soft}^T\}$ with \texttt{softargmax}.
	
	Considering a local $N\times N$ window, its scores are normalized with softmax:
	\begin{equation} \small
		\label{equ_ssoftmax}		
		s'(i,j) = \mathbf{softmax} \left( \frac{s(i,j)-s_{max}}{t_{det}} \right),
	\end{equation}
	where $t_{det}$ is the temperature which controls the ``sharpness'' of the normalization. And the $\mathbf{softmax}$ normalizes $\boldsymbol{x}$ as
	\begin{equation} \small
		\label{equ_softmax}
		\mathbf{softmax} (\boldsymbol{x}) = \frac{\exp (\boldsymbol{x}) }{\sum \exp (\boldsymbol{x}) }.
	\end{equation}
	The $s'(i,j)$ indicates the probabilities of $[i,j]^T$ to be the keypoint. Thus the expectation position of keypoint in the local window can be given by integral regression \cite{sun2018integral,gu2021removing}
	\begin{equation} \small
		\label{equ_softlocal}
		[\hat i, \hat j]_{soft}^T = \sum_{0\le i,j<N} s'(i,j)[i,j]^T.
	\end{equation}	
	Now, the output subpixel-level keypoints is given as
	\begin{equation} \small
		\label{equ_softglobal}
		\boldsymbol{p} = [u,v]_{soft}^T = [u,v]_{NMS}^T + [\hat i, \hat j]_{soft}^T .
	\end{equation}
	
	In this formulation, the first NMS term $[u,v]_{NMS}^T$ indicates the pixel-level keypoint position and is non-differentiable. While the second local soft coordinate term $[\hat i, \hat j]_{soft}^T$ represents an offset on the $[u,v]_{NMS}^T$ and is coupled with the scores in the local $N\times N$ window, making it differentiable in this window. Thus, the overall module is technically partially differentiable. In back-propagation, the gradient would flow to the scores in the local window through the second term, and thus optimizing the output keypoint position $[u,v]_{soft}^T$ is equivalent to optimizing the scores in the local window. Because there are many keypoints, the score map is sparsely optimized within individual local windows. This is similar to the sampling process in reinforced methods \cite{disk,bhowmik_reinforced_2020}, except that gradients can flow back into the scores in these windows.
	
	\begin{figure}[t]
		\centering
		\includegraphics[]{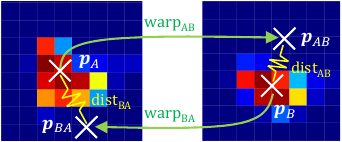}
		\caption{The reprojection loss pulls the warped keypoint ($\boldsymbol{p}_{AB}$, $\boldsymbol{p}_{BA}$) and its corresponding keypoint ($\boldsymbol{p}_{B}$, $\boldsymbol{p}_{A}$) together.
			In this figure, $\text{dist}_{AB}$ and $\text{dist}_{BA}$ are the reprojection distance between actual keypoints and warped keypoints, $\mathbf{warp}_{AB}$ and $\mathbf{warp}_{BA}$ are the differentiable warp functions.}
	\label{fig_reprojection}
\end{figure}
\subsection{Learning accurate keypoints}
\label{sec_kpt_rep_loss}
Accurate keypoints should be located precisely at repeatable locations (such as corners). For this purpose, we present the reprojection loss and dispersity peak loss. The reprojection loss directly optimizes the position of keypoints, and the dispersity peak loss ensures that the score is maximal at the keypoint position, resulting in accurate keypoints.

For two images $\boldsymbol{I}_A$ and $\boldsymbol{I}_B$, the network estimates their score maps $\boldsymbol{S}_A$ and $\boldsymbol{S}_B$, and the DKD module extracts the keypoints $\boldsymbol{p}_A$ and $\boldsymbol{p}_B$. Then $\boldsymbol{p}_A$ are warped to image $\boldsymbol{I}_B$ with
\begin{equation} \small
	\label{equ_warp}
	\boldsymbol{p}_{AB} = \mathbf{warp}_{AB}(\boldsymbol{p}_A),
\end{equation}
where $\mathbf{warp}_{AB}$ can be any differentiable warp function projecting keypoints from image A to image B, such as the homography projection, 3D perspective projection, or even the optical flow. The homography projection is given as
\begin{equation} \small
	[\boldsymbol{p}_{AB}^T, 1]^T = \boldsymbol{H}_{AB} [\boldsymbol{p}_A^T,1]^T,
\end{equation}
where $\boldsymbol{H}_{AB}$ denotes the $3\times 3$ homography matrix.
And the 3D perspective projection is
\begin{equation} \small
	\boldsymbol{p}_{AB} = \pi( d_A\boldsymbol{R}_{AB}\pi^{-1}(\boldsymbol{p}_A) + \boldsymbol{t}_{AB}),
\end{equation}
where $\pi(\boldsymbol{P}) = \boldsymbol{K}\boldsymbol{P}/Z$ projects a 3D point $\boldsymbol{P}=[X,Y,Z]^T$ in camera coordinate system to pixel coordinates on image plane with the camera intrinsics $\boldsymbol{K} \in \mathbb{R}^{2\times 3}$; $\boldsymbol{R}_{AB}$ and $\boldsymbol{t}_{AB}$ are the rotation and translation of 3D points from image A to B, respectively; and $d_A$ denotes the depth of keypoint $\boldsymbol{p}_A$.		

\subsubsection{Reprojection loss}
For a warped keypoint $\boldsymbol{p}_{AB}$, we find its closest detected keypoint $\boldsymbol{p}_{B}$ within $th_{gt}$ pixels distance as its corresponding keypoint. Then the reprojection distance of $\boldsymbol{p}_{AB}$ and $\boldsymbol{p}_{B}$ is defined as
\begin{equation} \small
	\label{equ_dist}
	\text{dist}_{AB} = \norm{ \boldsymbol{p}_{AB} - \boldsymbol{p}_{B} }_p,
\end{equation}
where $p$ is the norm factor. Inspired by symmetric epipolar distance, the reprojection loss is given in a symmetric way
\begin{equation} \small
	\label{equ_sydist}
	\begin{aligned}
		\mathcal{L}_{rp}                                   
		= \frac{ 1 } {2} (\text{dist}_{AB} + \text{dist}_{BA}).
	\end{aligned}
\end{equation}

As shown in Fig. \ref{fig_reprojection}, the minimization of reprojection loss pulls the warped keypoint and its corresponding keypoints together by adjusting their position. Since the keypoints are extracted from the score map using the differentiable DKD, the keypoint position is adjusted in a single optimization step by optimizing the scores within the local $N\times N$ window corresponding to individual keypoint. It implicitly provides keypoint repeatability \cite{repetability}, because the keypoints in non-repeatable areas would result in large reprojection errors. Therefore, other similarity measurements such as score differences \cite{keynet}, the cosine similarity \cite{r2d2} or the Kullbackleibler divergence \cite{sekd} are not necessary any more.

\begin{figure}[t]
	\centering
	\includegraphics[width=\linewidth]{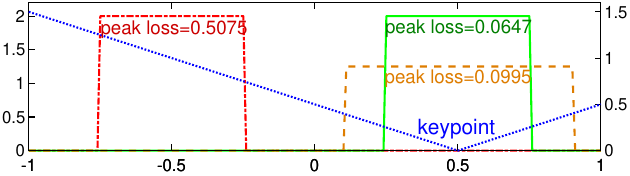}
	\caption{A toy example of one-denominational dispersity peak loss. In this example, the window range is $[-1,1]$, and the keypoint is at $x=0.5$. Three example rectangular score distributions are shown by the red dot-dash, green solid, and brown dash curves in the window. Their range is represented by the y-axis on the left. The blue dot curve is the distance to keypoint. Its range is shown on the right y-axis. The statistical characteristics (max-mean) of the red dot-dash and green solid distributions are identical, but they have different dispersity peak losses because of different spatial distribution.}
	\label{fig_peak}
\end{figure}

\subsubsection{Dispersity peak loss}
The minimization of reprojection loss optimizes the scores in the local window through the soft term $[\hat i, \hat j]_{soft}^T$ of equation \eqref{equ_softglobal}. However, the gradient step to improve $[\hat i, \hat j]_{soft}^T$ might affect the $[u,v]_{NMS}^T$. To align their optimization directions, we regularize the scores in the local window to be "peaky": that is, it should have a high score at the keypoint and low scores around it in the local window. In such a case, even if the local window centered on $[u,v]_{NMS}^T$ is slightly shifted, it still contains the keypoint, and $[\hat i, \hat j]_{soft}^T$ will be regulated to a new soft offset \textit{w.r.t.} the new local window center so that the keypoint position remains stable. In previous works \cite{r2d2,sekd}, this property is regularized by the difference between the maximum and average scores. However, it is only the statistical properties of the score patch, neglecting the spatial distribution of the scores. To force the score patch to be ``peaky'' exactly at the keypoint, we propose the score dispersity peak loss (Fig.\ref{fig_peak}). It takes the spatial distribution of scores into account, resulting in higher scores at the keypoint and lower scores further away.

Considering a $N\times N$ score patch, the distance of each pixel $[i,j]^T$ in the patch to the soft detected keypoint $[\hat i, \hat j]_{soft}^T$ is
\begin{equation} \small	
	d(i,j) = \left \{ \norm {[i,j] - [\hat i, \hat j]_{soft}} _p \ | \ 0\le i,j < N \right \}.
\end{equation}
The dispersity peak loss of this patch is then defined as 
\begin{equation} \small	
	\mathcal{L}_{pk} = \frac{1}{N^2} \sum_{0 \le i,j<N} d(i,j) s'(i,j),
\end{equation}
where $s'$ is the softmax score in equation (\ref{equ_softmax}).

\subsection{Learning discriminative descriptor}

\begin{figure}[t]
	\centering
	\includegraphics[]{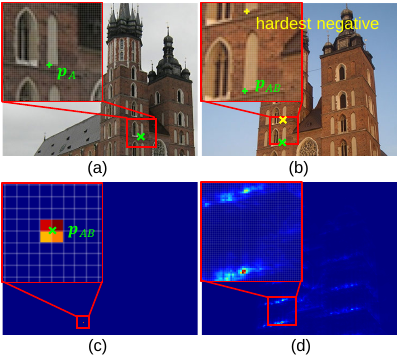}
	\caption{The illustration of triplet loss and neural reprojection loss. (a) the source image A, (b) the target image B, (c) the reprojection prabability map, (d) the matching probability map. A corresponding keypoint pair $\boldsymbol{p}_A$ and $\boldsymbol{p}_{AB}$ in image A and B are marked as green crosses. The yellow cross in image B is the position of the hardest negative descriptor of $\boldsymbol{p}_A$. The triplet loss of $\boldsymbol{p}_A$ is only constrained by $\boldsymbol{p}_{AB}$ and its hardest negative. While the NRE loss \cite{nre} minimizes the difference between the reprojection probability map and the matching probability map, connecting $\boldsymbol{p}_A$ to the dense descriptor map $\boldsymbol{D}_B$ and providing a more comprehensive constraint.}
	\label{fig_nre}
\end{figure}

Descriptors of the same keypoints (in different images) should be identical, whereas descriptors of different keypoints should be distinct. This is known as descriptor discriminativeness, and it is trained using the triplet loss \cite{tfeat,hardnet,sosnet,superpoint,sekd,d2net,aslfeat}. However, the triplet loss only optimizes the sparse descriptors sampled from keypoints, so the dense descriptor map cannot be fully constrained (For descriptor of $\boldsymbol{p}_{A}$ in Fig.\ref{fig_nre}(a), the triplet loss only uses the descriptors of two points in Fig.\ref{fig_nre}(b): the descriptors of the corresponding point $\boldsymbol{p}_{AB}$ and the hardest negative point.). To address this issue, we 
adopt the NRE loss \cite{nre} to train with the dense descriptor map. It minimizes the cross entropy difference between the dense reprojection probability map and the dense matching probability map (Fig. \ref{fig_nre}(c) and (d)), thereby providing a comprehensive constraint for the dense descriptor map as well as a stable training process (section \ref{sec_loss}).

As Fig. \ref{fig_nre}, for a keypoint $\boldsymbol{p}_{A}$ in image A, and its reprojected keypoint $\boldsymbol{p}_{AB}$ in image B, the reprojection probability map (Fig. \ref{fig_nre} (c)) is defined by $\boldsymbol{p}_{AB}$ with the bilinear interpolation:
\begin{equation} \small \small
	\begin{aligned}
		q_r(\boldsymbol{p}_B | \mathbf{warp}_{AB}, \boldsymbol{p}_A) := \ & w_{00} \llbracket \boldsymbol{p}_B = \left \lfloor \boldsymbol{p}_{AB} \right \rfloor \rrbracket +           \\
		& w_{01} \llbracket \boldsymbol{p}_B = \left \lfloor \boldsymbol{p}_{AB} \right \rfloor + [0,1]^T \rrbracket + \\
		& w_{10} \llbracket \boldsymbol{p}_B = \left \lfloor \boldsymbol{p}_{AB} \right \rfloor + [1,0]^T \rrbracket + \\
		& w_{11} \llbracket \boldsymbol{p}_B = \left \lfloor \boldsymbol{p}_{AB} \right \rfloor + [1,1]^T \rrbracket,
	\end{aligned}
\end{equation}
where $\llbracket \cdot \rrbracket$ is the Iverson bracket ($\llbracket True \rrbracket$=1 and $\llbracket False \rrbracket$=0), $\left \lfloor \cdot \right \rfloor$ denotes the floor function, and $w_{ij}$ are coefficients of bilinear interpolation. Following \cite{nre}, the inconsistent warped keypoints, \textit{e.g.} those outside the image, are marked as outlier ($\boldsymbol{out}$) with probability $q_r(\boldsymbol{p}_B | \mathbf{warp}_{AB}(\boldsymbol{p}_A)=\boldsymbol{out}) = 1$.

Besides, for the descriptor $\boldsymbol{d}_{\boldsymbol{p}_A} \in \mathbb{R}^{dim}$ of keypoint $\boldsymbol{p}_A$ and the dense descriptor map $\boldsymbol{D}_B\in \mathbb{R}^{(H\times W ) \times dim}$, their similarity map $\boldsymbol{C}_{\boldsymbol{d}_{\boldsymbol{p}_A},B} \in \mathbb{R}^{H\times W}$ is given as
\begin{equation} \small
	\label{equ_sim}
	\boldsymbol{C}_{\boldsymbol{d}_{\boldsymbol{p}_A},\boldsymbol{D}_B} = \boldsymbol{D}_B \boldsymbol{d}_{\boldsymbol{p}_A}.
\end{equation}
Again, an outlier category $\boldsymbol{out}$ is added to handle the case when the descriptor $\boldsymbol{d}_{\boldsymbol{p}_A}$ has no corresponding descriptor in the descriptor map $\boldsymbol{D}_B$. We denote the similarity map with outlier as $\overline {\boldsymbol{C}}_{\boldsymbol{d}_{\boldsymbol{p}_A},\boldsymbol{D}_B}:=\{ \boldsymbol{C}_{\boldsymbol{d}_{\boldsymbol{p}_A},\boldsymbol{D}_B}, \boldsymbol{out} \}$, thus $|\overline {\boldsymbol{C}}_{\boldsymbol{d}_{\boldsymbol{p}_A},\boldsymbol{D}_B}|=H\times W+1$. Now, the matching probability map (Fig. \ref{fig_nre} (d)) is given as the softmax normalization of the similarity map
\begin{equation} \small
	\label{equ_qm}
	q_m(\boldsymbol{p}_B|\boldsymbol{d}_{\boldsymbol{p}_A},\boldsymbol{D}_B) := \mathbf{softmax} \left( \frac{\overline {\boldsymbol{C}}_{\boldsymbol{d}_{\boldsymbol{p}_A},\boldsymbol{D}_B} -1} {t_{des}} \right),
\end{equation}
where $t_{des}$ controls the sharpness of matching probability map.

The NRE loss \cite{nre} minimizes the difference of the reprojection probability map and the matching probability map with cross-entropy (CE):
\begin{equation} \small
	\begin{aligned}
		& NRE(\boldsymbol{p}_A, \mathbf{warp}_{AB}, \boldsymbol{d}_{\boldsymbol{p}_A}, \boldsymbol{D}_B)\\
		& := CE\left(q_r(\boldsymbol{p}_B|\mathbf{warp}_{AB},\boldsymbol{p}_A) \| q_m(\boldsymbol{p}_B|\boldsymbol{d}_{\boldsymbol{p}_A},\boldsymbol{D}_B) \right)\\
		& = -\sum_{\boldsymbol{p}_B \in \{\boldsymbol{I}_B, \boldsymbol{out}\}} q_r(\boldsymbol{p}_B|\boldsymbol{p}_{AB}) \ln \left( q_m(\boldsymbol{p}_B|\boldsymbol{d}_{\boldsymbol{p}_A},\boldsymbol{D}_B) \right) \\
		& = - \ln \left( q_m(\boldsymbol{p}_{AB}| \boldsymbol{d}_{\boldsymbol{p}_A},\boldsymbol{D}_B) \right). \\
	\end{aligned}
\end{equation}
Hence, we define the descriptor loss in a symmetric way as 
\begin{equation} \small
	\begin{aligned}
		\mathcal{L}_{de} & = \frac{1}{N_A+N_B} * \\ 
		& (\sum_{\boldsymbol{p}_A \in \boldsymbol{I}_A} NRE(\boldsymbol{p}_A, \mathbf{warp}_{AB}, \boldsymbol{d}_{\boldsymbol{p}_A}, \boldsymbol{D}_B) + \\
		&\sum_{\boldsymbol{p}_B \in \boldsymbol{I}_B} NRE(\boldsymbol{p}_B, \mathbf{warp}_{BA}, \boldsymbol{d}_{\boldsymbol{p}_B}, \boldsymbol{D}_A) ),
	\end{aligned}
\end{equation}
where $N_A$ and $N_B$ are the number of keypoints in image A and B, respectively.

\subsection{Learning reliable keypoint}
The reprojection and dispersity peak loss provide accurate and repeatable keypoints. However, the spatial properties of descriptor map are not taken into account, so the keypoints might be unreliable \cite{r2d2}, \textit{e.g.}, the keypoints could locate in non-discriminative low-texture areas. To address this issue, we introduce a reliability loss based on the matching probability map in the NRE loss \cite{nre}.

First, the matching probability map is obtained by normalizing the similarity map $\boldsymbol{C}_{\boldsymbol{d}_{\boldsymbol{p}_A},\boldsymbol{D}_B} \in \mathbb{R}^{H\times W}$ in equation (\ref{equ_sim})
\begin{equation} \small
	\label{equ_c}
	\widetilde{\boldsymbol{C}}_{\boldsymbol{d}_{\boldsymbol{p}_A},\boldsymbol{D}_B} = \exp \left( \frac{\boldsymbol{C}_{\boldsymbol{d}_{\boldsymbol{p}_A},\boldsymbol{D}_B} - 1}{t_{rel}} \right),
\end{equation}
where $t_{rel}$ controls the sharpness. Then the reliability of keypoint $\boldsymbol{p}_A$ is defined as
\begin{equation} \small
	r_{\boldsymbol{p}_A} = \mathbf{bisampling}\left( \widetilde{\boldsymbol{C}}_{\boldsymbol{d}_{\boldsymbol{p}_A},\boldsymbol{D}_B}, \ \boldsymbol{p}_{AB} \right),
\end{equation}
where $\mathbf{bisampling}(\boldsymbol{M},\boldsymbol{p})$ is the bilinear sampling at position $\boldsymbol{p}\in \mathbb{R}^{2}$ on probability map $\boldsymbol{M} \in \mathbb{R}^{H\times W}$.

Intuitively, $r_{\boldsymbol{p}_A}$ assesses the matching quality of $\boldsymbol{p}_A$. 
If $\boldsymbol{p}_A$ is in unreliable low texture or repetitive region, the overall similarities in that region will be higher. As a result, the normalized similarity map $\widetilde{\boldsymbol{C}}_{\boldsymbol{d}_{\boldsymbol{p}_A},\boldsymbol{D}_B}$ has lower values, and the sampled score $r_{\boldsymbol{p}_A}$ is small, indicating that $\boldsymbol{p}_A$ is unreliable.

Considering all valid keypoints in image A, we define their reliability loss similar to D2Net \cite{d2net} and ASLFeat \cite{aslfeat} 
\begin{equation} \small
	\mathcal{L}_{reliability}^A = \frac{1}{N_A} \sum_{\substack{\boldsymbol{p}_A \in \boldsymbol{I}_A,\\ \boldsymbol{p}_{AB} \in \boldsymbol{I}_B }}  \frac{s_{\boldsymbol{p}_A}s_{\boldsymbol{p}_{AB}} }{\sum_{\substack{\boldsymbol{p}_A' \in \boldsymbol{I}_A,\\ \boldsymbol{p}_{AB}' \in \boldsymbol{I}_B}} s_{\boldsymbol{p}_A'}s_{\boldsymbol{p}_{AB}'}}(1-r_{\boldsymbol{p}_A}),
\end{equation}
where $N_A$ is the number of keypoints in the image A. For each keypoint $\boldsymbol{p}_A$ in image A, $\boldsymbol{p}_{AB}$ is its corresponding projection keypoint in image B (equation \eqref{equ_warp}). And $s_{\boldsymbol{p}}$ denotes the score value of keypoint $\boldsymbol{p}$ in its corresponding image. Similarly, the reliability loss is also given in a symmetric way as 
\begin{equation} \small
	\mathcal{L}_{rl} = \frac{1}{2}(\mathcal{L}_{reliability}^A + \mathcal{L}_{reliability}^B).
\end{equation}

\section{Experiments}
\label{sec_exp}
In this section, we first introduce the datasets, training details, and the evaluation metrics. To analyze the proposed method, ablation studies are conducted on network architecture and loss terms. At last, we reported the comparison results with the state-of-the-art methods on homography estimation, camera pose estimation, and visual (re-)localization tasks.

\subsection{Datasets}
\textbf{MegaDepth} \cite{megadepth} dataset includes tourist photos on famous sites and 3D maps built by COLMAP \cite{colmap}. It provides dense depth and camera pose for each image, which allow us to establish dense correspondences between images. We adopt the image pairs generated in DISK \cite{disk} to train our model. With co-visibility heuristics, it generates image pairs from the scenes except those overlapped with IMW2020 \cite{imw2020} validation and test sets, resulting 135 scenes with 63k images in total. 

\textbf{HPatches} \cite{hpatches} dataset contains planar images of 57 illumination and 59 viewpoint scenes. Each scene has 5 image pairs with ground truth homography matrices. Following D2Net \cite{d2net}, eight unreliable scenes are excluded. We conducted the ablation studies and homography estimation on this dataset.

\textbf{IMW2020} \cite{imw2020} is also built with tourist photos using COLMAP \cite{colmap}. It provides a standard pipeline for camera pose estimation, which we used to compare the proposed method to existing methods.

\textbf{Aachen Day-Night} \cite{aachen} dataset allows us to evaluate the effectiveness of descriptors on visual (re-)localization task. It tries to localize 98 query images captured in night-time based on a 3D model pre-built with day-time images.

\subsection{Training details}

\subsubsection{Details of loss calculation}

We used the DKD with a window size of $N=5$ to detect 400 keypoints  and randomly sampled another 400 keypoints on non-salient positions. In reprojection loss, the $th_{gt}=5$ and the normal factor $p=1$. The overall loss is
\begin{equation} \small
	\label{equ_overloss}
	\mathcal{L}= w_{rp}\mathcal{L}_{rp} + w_{pk}\mathcal{L}_{pk} + w_{rl}\mathcal{L}_{rl} + w_{de}\mathcal{L}_{de}.
\end{equation}
where $w_{rp}=1$, $w_{pk}=1$, $w_{rl}=1$, and $w_{de}=5$ in our experiments. And we set the normalization temperatures as $t_{det}=0.1$, $t_{rel}=1$, and $t_{des}=0.02$.

\subsubsection{Training setups}
The images were cropped and resized to $480\times 480$ in the training. The network was trained using the ADAM optimizer \cite{adam}, with the learning rate starting at zero and warming up to $3e^{-3}$ in 500 steps before remaining at $3e^{-3}$. We set the batch size to one, but accumulated the gradient over 16 batches. Under these settings, the proposed model converges on NVIDIA Titan RTX in about two days.

\begin{table}[t]
	\centering
	\caption{The network configurations. The ``$c_{i}$'' denotes the channel numbers of the $i$ block, and ``$N_{head}$'' is the number of layers in feature aggregation head. The ``MP'' is the number of parameters in millions. And the ``GFLOPs'' denotes the Giga FLoating-point OPerations of the network for $640\times480$ image.}
	\begin{tabular*}{\linewidth}{@{}@{\extracolsep{\fill}}lrrrrrrrr@{}}
		\hline
		\textit{Models} & $c_1$ & $c_2$ & $c_3$ & $c_4$ & $dim$ & $N_{head}$ & \textit{MP} & \textit{GFLOPs} \bigstrut \\ \hline
		\textbf{Tiny}   &     8 &    16 &    32 &    64 &    64 & 1 &       0.080 &           2.109 \bigstrut[t]\\
		\textbf{Small}  &     8 &    16 &    48 &    96 &    96 &              1&       0.142 &           3.893 \\
		\textbf{Normal} &    16 &    32 &    64 &   128 &   128 &              1&       0.318 &           7.909 \\
		\textbf{Large}  &    32 &    64 &   128 &   128 &   128 & 2 &       0.653 &          19.685 \bigstrut[b]\\ \hline
	\end{tabular*}%
	\label{tab_size}%
\end{table}%

\begin{table}[t]	
	\centering
	\caption{Ablation studies of network architectures.}
	\begin{tabular*}{\linewidth}{@{}@{\extracolsep{\fill}}lrrrr@{}}
		\hline
		\textit{Models}     &   $Rep$ &    $MS$ & $MMA@3$ &    $MHA@3$ \bigstrut \\ \hline
		\textbf{Tiny}      & 56.19\% & 32.54\% & 63.92\% & 70.56\% \bigstrut[t] \\
		\textbf{Small}     & 58.25\% & 34.96\% & 67.61\% &              73.52\% \\
		\textbf{Normal}    & 54.93\% & 38.58\% & 70.78\% &              75.74\% \\
		\textbf{Large}     & 53.70\% & 39.80\% & 70.50\% & 76.85\% \bigstrut[b] \\ \hline
	\end{tabular*}%
	\label{tab_backbone}%
\end{table}%

\subsection{Evaluation metrics}
Assuming that image A and image B have $N_A'$ and $N_B'$ co-visible keypoints, respectively, the number of co-visible keypoints is defined as $N_{cov}=(N_A'+N_B')/2$.
Among them, we get $N_{gt}$ ground truth keypoint pairs with the reprojection distance less than three pixels. On the other hand, $N_{putative}$ matches are obtained with mutual matching of the descriptors. And $N_{inlier}$ inlier matches are acquired by assessing the reprojection distance within different pixels threshold. Following previous works \cite{superpoint,hpatches}, we adopt the following metrics:

\begin{enumerate}
	\item \textbf{Repeatability} of keypoints is given as $Rep=N_{gt}/N_{cov}$. 
	
	\item \textbf{Matching Score} is given as $MS=N_{inlier}/N_{cov}$. 
	
	\item \textbf{Mean Matching Accuracy} ($MMA$) denotes the mean of matching accuracy of all image pairs, it is given as $N_{inlier}/N_{putative}$.
	
	\item \textbf{Mean Homography Accuracy} ($MHA$) is the mean of homograpy accuracy of all image pairs, it is defined as the percentage of correct image corners with the estimated homography matrix. 
	
\end{enumerate}

\subsection{Ablation Studies}
\label{sec_abl}

We conduct ablation studies on HPatches dataset \cite{hpatches} with single-scale. The score threshold $th$ in the DKD module is 0.2. Up to 5000 keypoints are detected. We evaluate the $MMA@3$ and $MHA@3$ without loss of generality.

\subsubsection{Ablation studies on network architecture}
Lightweight networks can improve running efficiency, but they have worse performance. To select a network that can run in real-time without significant performance degradation, we investigate four network variations (Table \ref{tab_size}). As in Table \ref{tab_backbone}, the matching score, matching accuracy, and homography accuracy of the ``Normal'' network increased by about 6\% compared with the ``Tiny'' network. However, despite doubling in complexity, the ``Large'' network only improves about 1\% compared with the ``Normal'' one, with diminishing marginal improvement. So, we use the ``Normal'' network to keep real-time performance.

\begin{table}[t]%
	\centering
	\caption{Ablation studies of losses. ``RP'', ``PK'', ``RL'', ``DE'', and ``Tri'' denote reprojection loss, dispersity peak loss, reliability loss, descriptor loss, and triplet loss, respectively.}
	\begin{tabularx}{\linewidth}{llllrrrr}
		\hline
		\textit{RP} & \textit{PK} & \textit{RL} & \textit{DE}  &            $Rep$ &             $MS$ &          $MMA@3$ &             $MHA@3$ \bigstrut \\ \hline
		&             & \checkmark  & \checkmark   &          59.26\% &          34.55\% &          71.83\% &          67.22\% \bigstrut[t] \\
		& \checkmark  & \checkmark  & \checkmark   &          48.75\% &          30.88\% &          62.95\% &                       72.59\% \\
		\checkmark  &             & \checkmark  & \checkmark   & \textbf{65.77\%} &          38.12\% & \textbf{74.44\%} &                       70.74\% \\
		\checkmark  & \checkmark  &             & \checkmark   &          51.11\% &          33.77\% &          65.23\% &                       74.44\% \\
		\checkmark  & \checkmark  & \checkmark  & \checkmark   &          54.93\% & \textbf{38.58\%} &          70.78\% & \textbf{75.74\%} \bigstrut[b] \\ \hline
		\checkmark  & \checkmark  & \checkmark  & \textit{Tri} &          57.91\% &          14.13\% &          39.38\% &             60.74\% \bigstrut \\ \hline
	\end{tabularx}%
	\label{tab_losses}%
\end{table}%

\begin{figure}[t]
	\centering
	\includegraphics[]{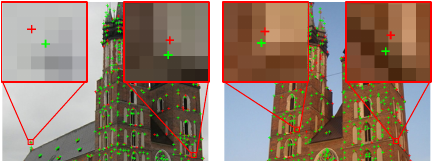}
	\caption{The effect of reprojection loss. The sub-pixel keypoints from the network trained with and without reprojection loss are denoted by green and red crosses, respectively.} 
\label{fig_abl_rep}
\end{figure}

\begin{figure}[t]
\centering
\includegraphics[scale=0.9]{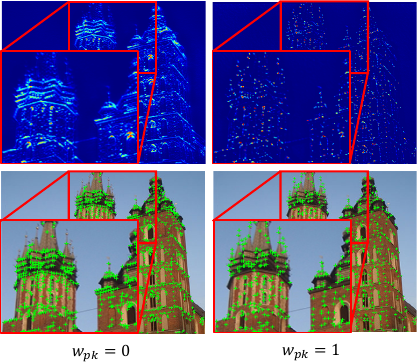}
\caption{The effect of dispersity peak loss. The score maps and keypoints detected from these score maps are shown in the first and second row.}
\label{fig_abl_pk}
\end{figure}

\begin{figure}[t]
\centering
\includegraphics[scale=0.9]{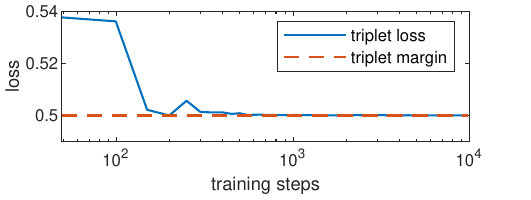}
\caption{The training process of the network with triplet loss.}
\label{fig_abl_desc}
\end{figure}

\subsubsection{Ablation studies on loss functions}
\label{sec_loss}
To investigate the loss functions, we train the ``Normal'' network with different loss configurations (Table \ref{tab_losses}).

\textbf{The reprojection loss} directly optimizes keypoints position and should produce accurate keypoints. In Fig. \ref{fig_abl_rep}, the visual comparisons of keypoints from networks trained with and without reprojection loss validates this anticipation: the keypoints of network trained with reprojection loss (green crosses) are more accurate than without reprojection loss (red crosses). Table \ref{tab_losses} shows the quantitative comparisons: the reprojection loss increases all metrics of the fifth row compared to the second, and the third row compared to the first.

\textbf{The dispersity peak loss} regularizes the shape of score patches. Table \ref{tab_losses} illustrates an intriguing phenomenon: the dispersity peak loss improves homography accuracy but decreases repeatability and matching accuracy (the second row compared to the first, and the fifth row compared to the third). To investigate this phenomenon, Fig. \ref{fig_abl_pk} visualizes the score map and keypoints  of network trained with and without the dispersity peak. With the dispersity peak loss, the score map becomes ``peaky'', which improves localization certainty and homography accuracy. While without the dispersity peak loss, the score map is less ``peaky'', keypoints are more likely to be crowded together. It increases the probability of finding a match in the crowd, so it is advantageous to repeatability and matching accuracy. As the accuracy of downstream tasks (homography estimation in this case) is more important, we retain the dispersity peak loss as the score map regularization.

\begin{table*}[htbp]
\centering
\caption{The number of network parameters, GFLOPs and inference FPS on $640\times 480$ images; the MMA and MHA within one to three pixels thresholds of different methods on Hpatches \cite{hpatches} dataset. The top three results are marked with \textcolor[rgb]{ 1,  0,  0}{\uuline{\textbf{red}}}, \textcolor[rgb]{ 0,  .62,  0}{\uline{\textbf{green}}}, and \textcolor[rgb]{0,0,1}{\uwave{\textbf{blue}}}.}

\begin{tabular*}{\textwidth}{@{}@{\extracolsep{\fill}}l|rrr|rrr|rrr@{}}
	\hline
	\textit{Models} & \textit{Params/M} & \textit{GFLOPs} & \textit{FPS} & \textit{MMA@1} & \textit{MMA@2} & \textit{MMA@3} & \textit{MHA@1} & \textit{MHA@2} & \textit{MHA@3} \bigstrut \\ \hline
	\textbf{D2-Net(MS)} \cite{d2net} & 7.635  & 889.40 & 6.60   & 9.78\% & 23.52\% & 37.29\% & 5.19\% & 21.30\% & 38.33\% \\
	\textbf{LF-Net(MS)} \cite{lfnet} & 2.642  & \textcolor[rgb]{ 0,  0,  1}{\uwave{\textbf{24.37}}} & 29.88  & 19.94\% & 41.98\% & 55.60\% & 17.41\% & 42.41\% & 57.78\% \\
	\textbf{SuperPoint} \cite{superpoint} & 1.301  & 26.11  & \textcolor[rgb]{ 0,  0,  1}{\uwave{\textbf{45.87}}} & 34.27\% & 54.94\% & 65.37\% & 35.00\% & 58.33\% & 70.19\% \\
	\textbf{R2D2(MS)} \cite{r2d2} & \textcolor[rgb]{ 0,  .62,  0}{\textbf{\uline{0.484}}} & 464.55 & 8.70   & 33.31\% & 62.17\% & \textcolor[rgb]{ 0,  .62,  0}{\textbf{\uline{75.77\%}}} & 35.74\% & 59.44\% & 71.48\% \\
	\textbf{ASLFeat(MS)} \cite{aslfeat} & 0.823  & 44.24  & 8.96   & 39.16\% & 61.07\% & 72.44\% & 37.22\% & 61.67\% & 73.52\% \\
	\textbf{DISK} \cite{disk} & 1.092  & 98.97  & 15.73  & 43.71\% & \textcolor[rgb]{ 1,  0,  0}{\textbf{\uuline{66.98\%}}} & \textcolor[rgb]{ 1,  0,  0}{\textbf{\uuline{77.59\%}}} & 34.07\% & 57.59\% & 70.56\% \bigstrut[b]\\
	\hline
	\textbf{ALIKE-N} & \textcolor[rgb]{ 1,  0,  0}{\textbf{\uuline{0.318}}} & \textcolor[rgb]{ 1,  0,  0}{\textbf{\uuline{7.91}}} & \textcolor[rgb]{ 1,  0,  0}{\textbf{\uuline{95.19}}} & 43.52\% & 63.14\% & 70.78\% & 42.04\% & \textcolor[rgb]{ 0,  0,  1}{\uwave{\textbf{62.78\%}}} & 75.74\% \bigstrut[t]\\
	\textbf{ALIKE-L} & \textcolor[rgb]{ 0,  0,  1}{\uwave{\textbf{0.653}}} & \textcolor[rgb]{ 0,  .62,  0}{\textbf{\uline{19.68}}} & \textcolor[rgb]{ 0,  .62,  0}{\textbf{\uline{68.18}}} & \textcolor[rgb]{ 0,  0,  1}{\uwave{\textbf{43.90\%}}} & 63.11\% & 70.50\% & \textcolor[rgb]{ 1,  0,  0}{\textbf{\uuline{45.00\%}}} & \textcolor[rgb]{ 1,  0,  0}{\textbf{\uuline{65.93\%}}} & \textcolor[rgb]{ 1,  0,  0}{\textbf{\uuline{76.85\%}}} \\
	\textbf{ALIKE-N(MS)} & \textcolor[rgb]{ 1,  0,  0}{\textbf{\uuline{0.318}}} & 25.97  & 41.48  & \textcolor[rgb]{ 0,  .62,  0}{\textbf{\uline{44.06\%}}} & \textcolor[rgb]{ 0,  0,  1}{\uwave{\textbf{65.56\%}}} & 74.05\% & \textcolor[rgb]{ 0,  0,  1}{\uwave{\textbf{44.07\%}}} & \textcolor[rgb]{ 0,  .62,  0}{\textbf{\uline{65.37\%}}} & \textcolor[rgb]{ 0,  0,  1}{\uwave{\textbf{75.93\%}}} \\
	\textbf{ALIKE-L(MS)} & \textcolor[rgb]{ 0,  0,  1}{\uwave{\textbf{0.653}}} & 64.63  & 29.15  & \textcolor[rgb]{ 1,  0,  0}{\textbf{\uuline{44.97\%}}} & \textcolor[rgb]{ 0,  .62,  0}{\textbf{\uline{66.21\%}}} & \textcolor[rgb]{ 0,  0,  1}{\uwave{\textbf{74.51\%}}} & \textcolor[rgb]{ 1,  0,  0}{\textbf{\uuline{45.00\%}}} & \textcolor[rgb]{ 0,  .62,  0}{\textbf{\uline{65.37\%}}} & \textcolor[rgb]{ 0,  .62,  0}{\textbf{\uline{76.48\%}}} \bigstrut[b]\\
	\hline
\end{tabular*}%
\label{tab_hpatches}%
\end{table*}%

\begin{table*}[htbp]	
\centering
\caption{Pose estimation results on IMW2020 \cite{imw2020} test sets (up to 2048 keypoints per image). The GFLOPs and performance per cost (PPC) are reported. For stereo task, we report Number of Features (NF), Rep, MS, mAA(5\degree) and mAA(10\degree). For multiview task, the Number of Matches (NM), Number of 3D Landmarks (NL), Track Length of landmark (TL), mAA(5\degree) and mAA(10\degree) are reported. The top three results are marked with \textcolor[rgb]{ 1,  0,  0}{\uuline{\textbf{red}}}, \textcolor[rgb]{ 0,  .62,  0}{\uline{\textbf{green}}}, and \textcolor[rgb]{0,0,1}{\uwave{\textbf{blue}}}.}
\begin{threeparttable}
	\setlength{\tabcolsep}{0.9mm}
	{
		\begin{tabular*}{\textwidth}{@{}@{\extracolsep{\fill}}l|r|rrrrrr|rrrrrr@{}}
			\hline
			\multicolumn{1}{c|}{\multirow{2}[2]{*}{\textit{\textbf{Methods}}}} & \multicolumn{1}{c|}{\multirow{2}[2]{*}{\textit{GFLOPs}}} & \multicolumn{6}{c|}{\textit{Stereo}}                & \multicolumn{6}{c}{\textit{Multiview}} \bigstrut[t]\\
			&        & \multicolumn{1}{c}{\textit{NF}} & \multicolumn{1}{c}{\textit{Rep}} & \multicolumn{1}{c}{\textit{MS}} & \multicolumn{1}{l}{\textit{mAA(5\degree)}} & \multicolumn{1}{l}{\textit{mAA(10\degree)}} & \multicolumn{1}{l|}{\textit{PPC}} & \multicolumn{1}{c}{\textit{NM}} & \multicolumn{1}{c}{\textit{NL}} & \multicolumn{1}{c}{\textit{TL}} & \multicolumn{1}{l}{\textit{mAA(5\degree)}} & \multicolumn{1}{l}{\textit{mAA(10\degree)}} & \multicolumn{1}{l}{\textit{PPC}} \bigstrut[b]\\
			\hline
			\textbf{D2-Net(MS)} \cite{d2net} & 889.40 & 2046   & 16.80\% & 29.30\% & 6.06\% & 12.27\% & 0.0138 & 2045.60 & 1999.37 & 3.01   & 17.77\% & 28.30\% & 0.0318 \bigstrut[t]\\
			\textbf{LF-Net(MS)}\tnote{*} \cite{lfnet} & \textcolor[rgb]{ 0,  0,  1}{\uwave{\textbf{24.37}}} &  ---   &  ---   &  ---   &  ---   & 23.44\% & 0.9620 & 196.70 & 1385.00 & 4.14   &  ---   & 51.41\% & \textcolor[rgb]{ 0,  0,  1}{\uwave{\textbf{\uline{2}.1099}}} \\
			\textbf{SuperPoint} \cite{superpoint} & 26.11  & 2048   & 36.40\% & 63.00\% & 19.71\% & 28.97\% & \textcolor[rgb]{ 0,  0,  1}{\uwave{\textbf{1.1093}}} & 2048.00 & 1185.38 & 4.33   & 44.35\% & 54.66\% & 2.0935 \\
			\textbf{R2D2(MS)} \cite{r2d2} & 464.55 & 2048   & 42.90\% & 74.60\% & 27.20\% & 39.02\% & 0.0840 & 2048.00 & 1225.85 & 4.28   & 53.13\% & 64.03\% & 0.1378 \\
			\textbf{ASLFeat(MS)} \cite{aslfeat} & 77.58  & 2043   & \textcolor[rgb]{ 0,  0,  1}{\uwave{\textbf{43.10\%}}} & 74.90\% & 22.62\% & 33.65\% & 0.4337 & 157.51 & 1106.59 & 4.42   & 45.28\% & 55.61\% & 0.7168 \\
			\textbf{DISK} \cite{disk} & 98.97  & 2048   & \textcolor[rgb]{ 1,  0,  0}{\textbf{\uuline{44.80\%}}} & \textcolor[rgb]{ 1,  0,  0}{\textbf{\uuline{85.20\%}}} & \textcolor[rgb]{ 1,  0,  0}{\textbf{\uuline{38.72\%}}} & \textcolor[rgb]{ 1,  0,  0}{\textbf{\uuline{51.22\%}}} & 0.5175 & 526.35 & 2424.80 & 5.50   & \textcolor[rgb]{ 1,  0,  0}{\textbf{\uuline{63.25\%}}} & \textcolor[rgb]{ 1,  0,  0}{\textbf{\uuline{72.96\%}}} & 0.7372 \\
			\textbf{ALike-N} & \textcolor[rgb]{ 1,  0,  0}{\textbf{\uuline{7.91}}} & 1803   & \textcolor[rgb]{ 0,  .62,  0}{\textbf{\uline{43.30\%}}} & \textcolor[rgb]{ 0,  0,  1}{\uwave{\textbf{81.10\%}}} & \textcolor[rgb]{ 0,  0,  1}{\uwave{\textbf{35.12\%}}} & \textcolor[rgb]{ 0,  0,  1}{\uwave{\textbf{47.18\%}}} & \textcolor[rgb]{ 1,  0,  0}{\textbf{\uuline{5.9652}}} & 276.48 & 1644.20 & 4.97   & \textcolor[rgb]{ 0,  0,  1}{\uwave{\textbf{59.18\%}}} & \textcolor[rgb]{ 0,  0,  1}{\uwave{\textbf{69.21\%}}} & \textcolor[rgb]{ 1,  0,  0}{\textbf{\uuline{8.7507}}} \\
			\textbf{ALike-L} & \textcolor[rgb]{ 0,  .62,  0}{\textbf{\uline{19.68}}} & 1771   & 42.90\% & \textcolor[rgb]{ 0,  .62,  0}{\textbf{\uline{82.20\%}}} & \textcolor[rgb]{ 0,  .62,  0}{\textbf{\uline{37.24\%}}} & \textcolor[rgb]{ 0,  .62,  0}{\textbf{\uline{49.58\%}}} & \textcolor[rgb]{ 0,  .62,  0}{\textbf{\uline{2.5187}}} & 298.30 & 1693.31 & 5.02   & \textcolor[rgb]{ 0,  .62,  0}{\textbf{\uline{60.30\%}}} & \textcolor[rgb]{ 0,  .62,  0}{\textbf{\uline{70.22\%}}} & \textcolor[rgb]{ 0,  .62,  0}{\textbf{\uline{3.5673}}} \bigstrut[b]\\
			\hline
		\end{tabular*}%
	}
	\begin{tablenotes}
		\item[*] There are no public test results of LF-Net \cite{lfnet} on the benchmark website, and the test results came from DISK \cite{disk}.
	\end{tablenotes}
\end{threeparttable}
\label{tab_imw2020}%
\end{table*}%

\textbf{The descriptor loss} were studied by comparing the NRE loss \cite{nre} and triplet loss \cite{tfeat}. For triplet loss, negative descriptor is mined from the keypoints (Fig. \ref{fig_nre}(b)), and the triplet margin is 0.5. Fig. \ref{fig_abl_desc} shows the training process of triplet loss with sub-pixel keypoints. It falls into a local minimum where the loss approaches but dose not cross the triplet margin (all the descriptors become the same). So its performance are extremely worse (the last row of Table \ref{tab_losses}). We find it is tricky to tune the hyperpameters for triplet loss, so we adopt the NRE loss \cite{nre} as it has a stable convergence, although it requires more GPU memory and training time because of dense similarity map computation in equation \eqref{equ_sim}.

\textbf{The reliability loss} ensures that the keypoints are located at areas where the descriptors are discriminative. More reliable keypoints would produce fewer false matches, resulting in a higher matching score and accuracy, as shown in the fourth and fifth row of Table \ref{tab_losses}.

\begin{figure*}[htbp]
\centering
\includegraphics[]{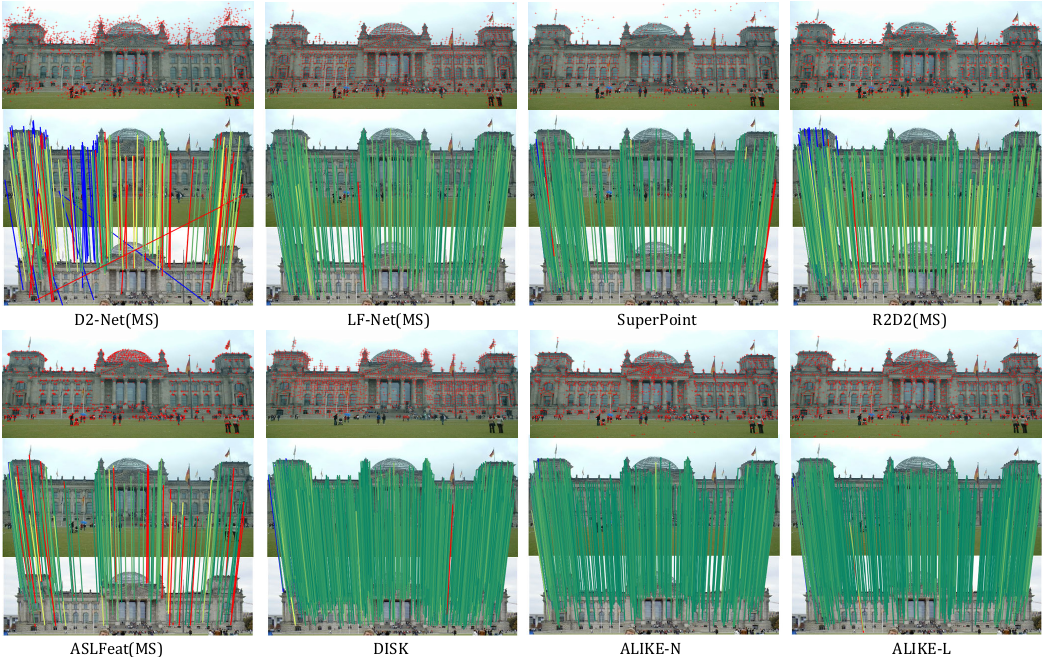}
\caption{The visualization of detected keypoints (the first and third row) and the matches (the second and fourth row) on IMW2020 \cite{imw2020} validation set. In the the second and fourth row, the inliers are plotted from green to yellow if they are correct (0 to 5 pixels in reprojection error), in red if they are incorrect (above 5 pixels), and in blue if ground truth depth is not available. Best viewed in color and zoom in 400\%.}
\label{fig_imw}
\end{figure*}

\subsection{Comparisons with the state-of-the-arts}
The proposed methods are compared to state-of-the-arts on homography estimation, camera pose estimation, and visual (re-)localization tasks. And we denote the proposed methods as ``ALIKE-[T/S/N/L][(MS)]'', where ``[T/S/N/L]'' is model size (Table \ref{tab_size}), and ``MS'' is multi-scale keypoint detection. 

\subsubsection{The network complexity}
Table \ref{tab_hpatches} reports the complexity of different methods, including the number of parameters, GFLOPs (Giga FLoating-point OPerations), and the inference FPS. As can be seen, the ALIKE-N has the fewest parameters (318K), followed by R2D2 \cite{r2d2} (484K) and ALIKE-L (653K). However, a network with fewer parameters does not mean it has less computation due to non-parametric operations. For example, R2D2 \cite{r2d2} has 464.55 GFLOPs computational cost despite having only 484K parameters. While ALIKE-N (318K) and ALIKE-L (653K) only have it for 7.91 and 19.68 GFLOPs, respectively. For more intuitive comparisons, the inference FPS are also reported in Table \ref{tab_hpatches}. The ALIKE-N, ALIKE-L, and SuperPoint \cite{superpoint} are the fastest, can run at 95.19, 68.18, and 45.87 FPS, respectively.  Considering the matching performance, the proposed methods have a short inference time while also provide accurate transformation estimation. 

\subsubsection{Homography estimation}
\label{sec_homo}
Following previous works \cite{d2net,r2d2,aslfeat}, we conduct homography estimation on HPatches dataset \cite{hpatches}. Previous works focus on the matching accuracy (MMA), but we believe it is only an intermediate metric, as the goal of keypoints matching is downstream homography estimation. For better homography estimation accuracy, the MMA at stricter thresholds and the homography accuracy (MHA) is more important. Therefore, we report MMA and MHA at stricter thresholds in Table \ref{tab_hpatches}.
Due to the DKD module and proposed losses, the proposed methods have much better MMA at stricter thresholds than previous works. This indicates that accurate keypoints are obtained with the proposed method. More importantly, for homography accuracy MHA, the proposed methods outperform SOTA methods.
Compared to ASLFeat(MS) \cite{aslfeat}, the method with the highest MHA@3, the proposed ALIKE-N and ALIKE-L improve MHA@3 by 2.22\% and 3.33\%, respectively, while decreases computational complexity by about 5.6 and 2.2 times.

\subsubsection{Camera pose estimation}
The IMW2020 \footnote{\url{https://www.cs.ubc.ca/research/image-matching-challenge/}} \cite{imw2020} includes the stereo and multi-view tasks for pose estimation. For fair comparison, we used the best configuration for each method and detected up to 2048 key points using the built-in mutual nearest neighbor matching. The mean Average Accuracy (mAA) is obtained by integrating the  translation and rotation vector errors to 5\degree \ and 10\degree. As real-time performance is also important for practical applications, the GFLOPs and performance per cost (PPC) \cite{VPR-bench}, where $PPC=mAA(10\degree)/GFLOPs$, are also reported in Table \ref{tab_imw2020}. 

Table \ref{tab_imw2020} illustrates the quantitative evaluation results. Without considering the PPC, the proposed ALIKE outperforms all previous methods except DISK \cite{disk}. However, DISK requires 98.97 GFLOPs of computational cost, whereas the ALIKE-L and ALIKE-N require only 7.91 and 19.68 GFLOPs, respectively. Considering the PPC, the ALIKE-N and ALIKE-L are approximately 10 and 5 times more efficient than DISK, respectively. In summary, the proposed ALIKE is both accurate and lightweight, making it ideal for real-time applications.

To examine each method, the detected keypoints and estimated matches are visualized in Fig. \ref{fig_imw}. The D2-Net \cite{d2net} and ASLFeat \cite{aslfeat} extract keypoints from low-resolution feature maps. So their keypoints are less accurate, resulting in some error matches (the red lines). For LF-Net \cite{lfnet}, R2D2 \cite{r2d2}, and DISK \cite{disk}, many false keypoints are in the texture-less building boundary. Keypoints of R2D2 \cite{r2d2} tend to crammed together, whereas keypoints of DISK \cite{disk} are almost all on buildings. SuperPoint \cite{superpoint} generates the sparsest keypoints, and some of them are in unreliable sky. While the majority keypoints of the proposed methods are located at image corners and edges, and less incorrect matches are included.

\begin{table}[t]
\centering
\caption{Visual localization results on Aachen Day-night \cite{aachen} dataset. Percentages of localized queries within three tolerances using top-2048 and unlimited keypoints are reported. The top three results are marked as \textcolor[rgb]{ 1,  0,  0}{\uuline{\textbf{red}}}, \textcolor[rgb]{ 0,  .62,  0}{\uline{\textbf{green}}}, and \textcolor[rgb]{0,0,1}{\uwave{\textbf{blue}}}.}
\setlength{\tabcolsep}{0.5mm}
{
	\begin{tabular*}{\linewidth}{@{}@{\extracolsep{\fill}}l|rrr|rrr@{}}
		\hline
		\multicolumn{1}{c|}{\multirow{2}[2]{*}{\textbf{Methods}}} &                                                                \multicolumn{3}{c|}{2048 keypoints}                                                                 &                                                              \multicolumn{3}{c}{unlimited keypoints} \bigstrut[t]                                                               \\
		&                  \multicolumn{1}{l}{ 0.25m,2\degree} &                    \multicolumn{1}{l}{0.5m,5\degree} &                    \multicolumn{1}{l|}{5m,10\degree} &                  \multicolumn{1}{l}{ 0.25m,2\degree} &                    \multicolumn{1}{l}{0.5m,5\degree} &                     \multicolumn{1}{l}{5m,10\degree} \bigstrut[b] \\ \hline
		\textbf{D2-Net(SS)} \cite{d2net}                                   &  \textcolor[rgb]{ 1,  0,  0}{\textbf{\uuline{74.5}}} & \textcolor[rgb]{ 0,  .62,  0}{\textbf{\uline{85.7}}} & \textcolor[rgb]{ 0,  .62,  0}{\textbf{\uline{96.9}}} &   \textcolor[rgb]{ 0,  0,  1}{\uwave{\textbf{78.6}}} & \textcolor[rgb]{ 0,  .62,  0}{\textbf{\uline{88.8}}} & \textcolor[rgb]{ 1,  0,  0}{\textbf{\uuline{100.0}}} \bigstrut[t] \\
		\textbf{D2-Net(MS)} \cite{d2net}                                   &                                                 61.2 &                                                 81.6 &   \textcolor[rgb]{ 0,  0,  1}{\uwave{\textbf{94.9}}} & \textcolor[rgb]{ 0,  .62,  0}{\textbf{\uline{79.6}}} &                                                 86.7 &              \textcolor[rgb]{ 1,  0,  0}{\textbf{\uuline{100.0}}} \\
		\textbf{SEKD(SS)} \cite{lfnet}                                     &                                                 42.9 &                                                 51.0 &                                                 57.1 &                                                 54.1 &                                                 65.3 &                                                              74.5 \\
		\textbf{SEKD(MS)} \cite{lfnet}                                     &                                                 50.0 &                                                 63.3 &                                                 70.4 &                                                 59.2 &                                                 72.4 &                                                              82.7 \\
		\textbf{SuperPoint} \cite{superpoint}                              &   \textcolor[rgb]{ 0,  .62,  0}{\uline{\textbf{72.4}}} &                                                 79.6 &                                                 88.8 &                                                 73.5 &                                                 81.6 &                \textcolor[rgb]{ 0,  0,  1}{\uwave{\textbf{93.9}}} \\
		\textbf{R2D2(MS)} \cite{r2d2}                                      &                                                 63.3 &                                                 78.6 &                                                 87.8 &                                                 71.4 &                                                 85.7 &              \textcolor[rgb]{ 0,  .62,  0}{\textbf{\uline{99.0}}} \\
		\textbf{ASLFeat(SS)} \cite{aslfeat}                                &                                                 54.1 &                                                 67.3 &                                                 76.5 &                                                 77.6 & \textcolor[rgb]{ 0,  .62,  0}{\textbf{\uline{88.8}}} &              \textcolor[rgb]{ 1,  0,  0}{\textbf{\uuline{100.0}}} \\
		\textbf{ASLFeat(MS)} \cite{aslfeat}                                &                                                 49.0 &                                                 59.2 &                                                 69.4 &                                                 77.6 &  \textcolor[rgb]{ 1,  0,  0}{\textbf{\uuline{90.8}}} &              \textcolor[rgb]{ 1,  0,  0}{\textbf{\uuline{100.0}}} \\
		\textbf{DISK} \cite{disk}                                          & \textcolor[rgb]{ 0,  0,  1}{\textbf{\uwave{70.4}}} &                                                 82.7 &   \textcolor[rgb]{ 0,  0,  1}{\uwave{\textbf{94.9}}} &  \textcolor[rgb]{ 1,  0,  0}{\textbf{\uuline{84.7}}} &  \textcolor[rgb]{ 1,  0,  0}{\textbf{\uuline{90.8}}} &              \textcolor[rgb]{ 1,  0,  0}{\textbf{\uuline{100.0}}} \\
		\textbf{ALIKE-N}                                         &                                                 68.4 &   \textcolor[rgb]{ 0,  0,  1}{\uwave{\textbf{84.7}}} & \textcolor[rgb]{ 0,  .62,  0}{\textbf{\uline{96.9}}} &                                                 77.6 &   \textcolor[rgb]{ 0,  0,  1}{\uwave{\textbf{87.8}}} &              \textcolor[rgb]{ 1,  0,  0}{\textbf{\uuline{100.0}}} \\
		\textbf{ALIKE-L}                                         &  \textcolor[rgb]{ 1,  0,  0}{\textbf{\uuline{74.5}}} &  \textcolor[rgb]{ 1,  0,  0}{\textbf{\uuline{87.8}}} &  \textcolor[rgb]{ 1,  0,  0}{\textbf{\uuline{98.0}}} & \textcolor[rgb]{ 0,  .62,  0}{\textbf{\uline{79.6}}} &                                                 86.7 & \textcolor[rgb]{ 1,  0,  0}{\textbf{\uuline{100.0}}} \bigstrut[b] \\ \hline
	\end{tabular*}%
}
\label{tab_aachen}%
\end{table}%

\subsubsection{Visual (re-)localization}
The proposed method was tested on Aachen Day-Night visual (re-)localization benchmark\footnote{\url{https://www.visuallocalization.net}} \cite{aachen} with default configurations. It builds a 3D map based on image keypoints, registers query images to the map, and uses the percentage of correctly registered images in three error tolerances (\textit{i.e.} ($0.25m, 2\degree$)/($0.5m,5\degree$)/($5m,10\degree$)) as evaluation metric. We report performance when using limited (up to 2048) and unlimited features in Table \ref{tab_aachen}.

Except SEKD \cite{sekd}, most methods perform well when using unlimited number of keypoints. But practical applications typically use limited number of keypoints, D2-Net(MS) \cite{d2net}, ASLFeat \cite{aslfeat}, and DISK \cite{disk} degrade dramatically in such cases. ALIKE-L achieves superior performance with fewer keypoints even without heuristically fine-tuning configurations, demonstrating the required effectiveness in resource-constrained applications. Furthermore, the normal model (ALIKE-N) achieves comparable localization accuracy while using fewer computational resources.

\subsection{Limitations of proposed method}
\begin{figure}[htbp]
\centering
\includegraphics[width=\linewidth]{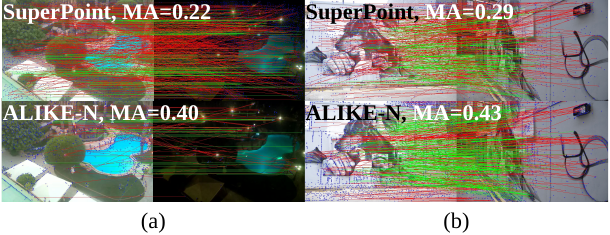}
\caption{The visualization of matching results of SuperPoint \cite{superpoint} and ALIKE-N on (a) challenging illumination and (b) viewpoint image pairs. The keypoints are represented by blue crosses, while correct matches (with a reprojection error $<$ 3 pixels) and incorrect matches are represented by green and red lines, respectively. In this figure, only the top-500 putative matches with the highest similarity are shown for clarity.}
\label{fig_limit}
\end{figure}

We observed two kinds of failure cases in image matching using the proposed keypoint and descriptor: images with extreme illumination changes and large viewpoint differences (as shown in Fig. \ref{fig_limit}), which are also the two most difficult challenges in image matching task. Technically, our lightweight network utilizes shallow architecture (Table \ref{tab_size}) and will be limited to extracting more low-level descriptors, which sacrifices some presentation capabilities. As a consequence, matching performance in extreme scenarios will be challenging. However, in the randomly selected challenging cases, the matching accuracy (MA, number of correct matches / number of putative matches) of ALIKE-N is twice as high as SuperPoint \cite{superpoint}, which has the highest FPS and PPC except ours but three times the GFLOPs of ALIKE-N (Fig. \ref{fig_limit}).

\section{Conclusions}
\label{sec_con}
In this paper, we present the ALIKE, an end-to-end accurate and lightweight keypoint detection and descriptors extraction network. It uses the differentiable keypoint detection module to detect accurate sub-pixel keypoints. And the keypoints are then trained with proposed reprojection loss and dispersity peak loss. Besides the keypoints, the NRE loss are used to train discriminative descriptors and the reliability loss are presented to force reliable keypoints. Compared to state-of-the-art approaches, the proposed method achieves comparable or superior results on homography estimation, camera pose estimation, and visual (re-)localization tasks while taking significantly less time to run. The ALIKE-N and ALIKE-L can run at 95 FPS and 68 FPS, respectively, for $640\times 480$ images on NVIDIA TITAN X (Pascal). Our future works include enhancing the reliability and accuracy of detected keypoints using high-level semantic information or reinforcement learning, as well as integrating the proposed model into practical SLAM and HDR applications.

\normalem
\bibliographystyle{IEEEtran}


\end{document}